\documentclass[letterpaper, 10 pt, conference]{ieeeconf}
\IEEEoverridecommandlockouts
\usepackage{cite}
\usepackage{amsmath,amssymb,amsfonts}
\usepackage{graphicx}
\usepackage{xspace}
\usepackage{pifont}
\usepackage{caption}
\usepackage{tikz}

\def\BibTeX{{\rm B\kern-.05em{\sc i\kern-.025em b}\kern-.08em
    T\kern-.1667em\lower.7ex\hbox{E}\kern-.125emX}}

\newcommand{\highwayenv}{highway-env\xspace}
\newcommand{\nuscenes}{nuScenes\xspace}

\newcommand{\cmark}{\ding{51}}
\newcommand{\xmark}{\ding{55}}

\begin{document}

\title{Scaling Planning for Automated Driving using Simplistic Synthetic Data}

\author{ Martin~Stoll,~Markus~Mazzola,~Maxim~Dolgov,~J\"{u}rgen~Mathes,~and~Nicolas~M\"{o}ser$^\dag$
\thanks{$^\dag$The authors are with the Robert Bosch GmbH, Corporate Research, D-71272 Renningen, Germany. E-Mail: {\tt\small\{martin.stoll, markus.mazzola, maxim.dolgov, juergen.mathes, nicolas.moeser\}@de.bosch.com}}
}

\maketitle

\begin{abstract}
We challenge the perceived consensus that the application of deep learning to solve the automated driving planning task necessarily requires huge amounts of real-world data or highly realistic simulation.
Focusing on a roundabout scenario, we show that this requirement can be relaxed in favour of targeted, simplistic simulated data.
A benefit is that such data can be easily generated for critical scenarios that are typically underrepresented in realistic datasets.
By applying vanilla behavioural cloning almost exclusively to lightweight simulated data, we achieve reliable and comfortable driving in a real-world test vehicle.
We leverage an incremental development approach that includes regular in-vehicle testing to identify sim-to-real gaps, targeted data augmentation, and training scenario variations.
In addition to a detailed description of the methodology, we share our lessons learned,
touching upon scenario generation, simulation features, and evaluation metrics.
\end{abstract}

\section{Introduction}
\label{introduction}

Decision making in automated driving is an unsolved open-world problem.
Especially the unknown intents and behaviour strategies of other agents pose a challenge.
Engineered approaches rely on heuristics that cannot cover the variety of the behaviour strategies of other agents.
Thus, to be able to deliver safe driving, planners are often optimised toward an over-cautious and passive behaviour.
This limitation harms performance and can lead to the "frozen robot problem"~\cite{trautman2010frozenrobot}.
Additionally, engineering heuristics usually does not scale to different operational domains and prevents the dissimination of automated driving.
As an alternative, using learned models in existing planners or even learning the entire planner has emerged.

Early generations of learned planners focused on vision-based end-to-end planning where the sensor data was directly mapped to low-level actions such as acceleration and steering angle~\cite{bojarski2016endtoend, codevilla2018end, bewley2018driveinaday}.
However, an alternative representation -- a birds-eye-view grid with waypoints or trajectory output -- quickly established as a baseline and led to many approaches with a Convolutionaly Neural Network (CNN) backbone for planning and prediction such as~\cite{Bansal2018ChauffeurNet, hong2019rules, cui2019multimodal, chai2019multipath}.
The usage of a grid-based input representation even allowed to learn the full pipeline comprising perception, prediction, and planning with intermittent interpretable interfaces~\cite{zeng2019end, sadat2020perceive, casas2021mp3, cui2021lookout}.
Recently, graph-based representations and attention-based backbones started to gain momentum and produce a large body of research, e.g.~\cite{liang2020lanegcn, bhat2020trajformer}.

Unfortunately, prior work has demonstrated that training a planner requires large amounts of demonstration data.
For example, \cite{Bansal2018ChauffeurNet} reported using a dataset containing 1440 hours while \cite{zeng2019end} was trained with approximately 40  hours of driving.
However, even if a large dataset is available or different datasets such as~\cite{caesar2020nuscenes, ettinger2021large, caesar2021nuplan} are combined, pure imitation is not sufficient because challenging situations are highly underrepresented~\cite{Bansal2018ChauffeurNet}.
This challenge can be addressed for example by augmenting demonstration data with perturbations~\cite{Bansal2018ChauffeurNet}, via dataset augmentation methods like DAgger~\cite{ross2011dagger}, or by using differentiable simulation~\cite{scheel2022urban}.
However, these methods have been demonstrated only on small-scale variations of the scenes from the dataset, e.g.\ by small variations of relative distances or velocities between the agents.
In this way, a certain robustness margin for the scenes in the dataset can be learned but these variations do not create entirely new or underrepresented scenarios.

\begin{figure}
	\centering
	\begin{tikzpicture}
		[every node/.style={align=center,anchor=base},baseline=0pt]
		\node[inner sep=20pt] (model) at (0,0.5) {\includegraphics[width=.4\textwidth]{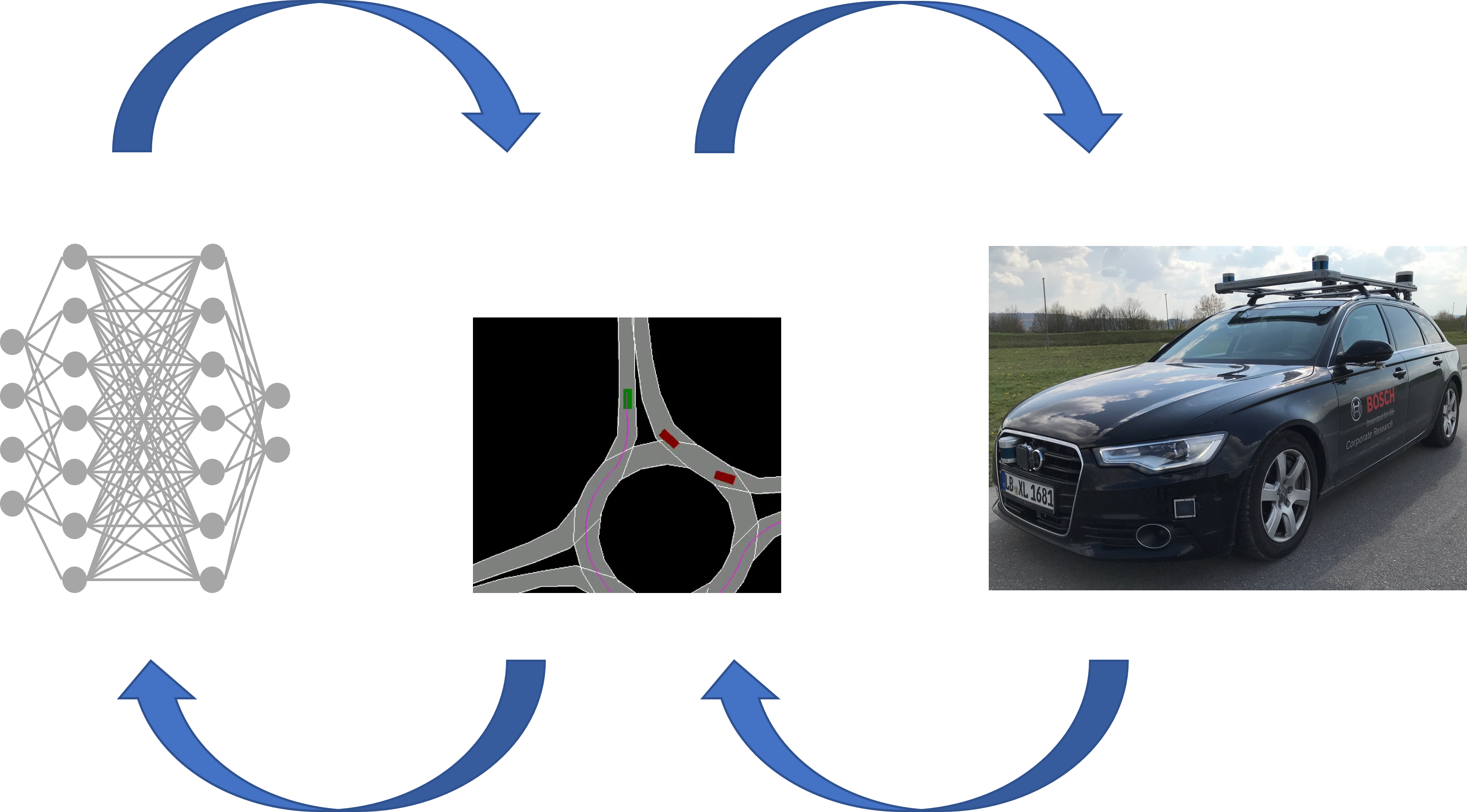}};
		\node[text width=60pt] (input) at (-2.1,5) {Benchmark on simulation};
		\node[text width=60pt] (input) at (0.8,5) {Observe sim2real gap};
		\node[text width=60pt] (input) at (-0.5,3.45) {Simplistic simulation};
		\node[text width=70pt] (input) at (-2.0,0) {Train with newly simulated data};
		\node[text width=60pt] (input) at (0.9,0) {Extend simulation};
	\end{tikzpicture}
	\caption{Alternating steps of our itertive development cycle: train and optimise planner with simplistic simulation (left), observe sim2real gap in on-road test and extend simulator (right).}
	\label{fig:page1}
\end{figure}
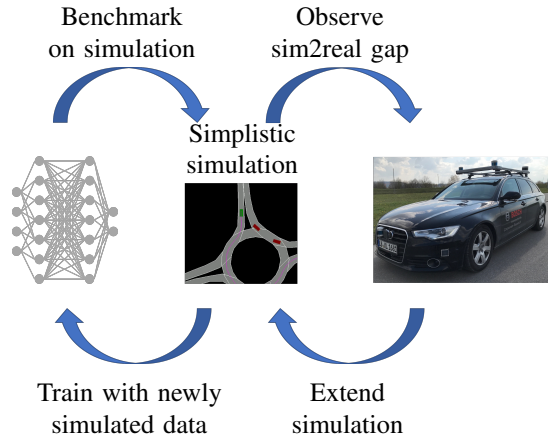

In this paper, we address this gap by creating entirely new scenarios in a lightweight simulation, thus extending the training data with large-scale variations (also see Fig.~\ref{fig:page1}).
We show that even applying vanilla behavioural cloning almost exclusively on lightweight simulated data can enable safe and comfortable driving in the real world and can generalise to scenarios trained on simulated data alone.
We leverage the fact that implementing a simple but performant expert for a specific scenario is much easier than creating a policy that generalises to the plethora of real-world situations.
Our approach also circumvents the need to develop meaningful and realistic driver models for other agents~\cite{suo2021trafficsim, igl2022symphony}, which would be necessary e.g.\ for reinforcement learning or randomised scenario mining.
Additionally, it allows to infuse expert knowledge into the selection of challenging scnenarios to cover a large portion of the open-context planning problem.
With this approach, we obtain training data for challenging scenarios and road topologies that are not present in the available real-world demonstration data.

When generating synthetic data using simulation, there is always a varying degree of difference between simulated and real-world distributions, known as sim2real gap. Such a mismatch can be introduced at all stages of the simulation process, e.g.\ modelling of the environment, perception, the ego vehicle itself, or behaviour of other traffic participants.
The obvious solution, making the simulation more realistic, is generally a difficult task and not always feasible.

A key aspect of our approach is not the planning model but rather an iterative development cycle. We identify shortcomings of a model that is trained on a version of our mixed dataset in simulation and then (after passing simulation quality gates) test the model in the car. Persisting or newly identified problems are reproduced in simulation. Whereever possible, we fix effects caused by sim2real gaps by using data augmentations. Shortcomings of the model itself are fixed by either using data augmentations or by generating targeted scenarios in our simplistic simulation.

\section{Model Description}
\label{model}
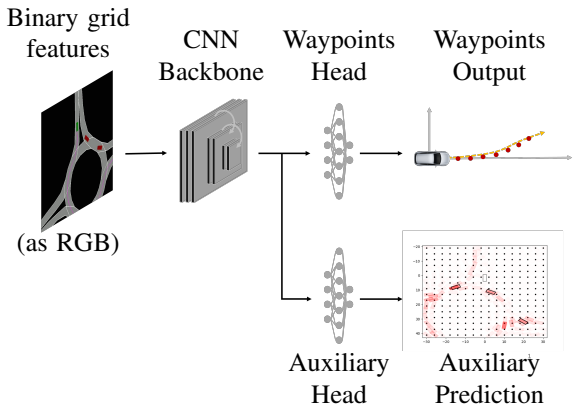
\begin{figure}
	\centering
	\begin{tikzpicture}
		[every node/.style={align=center,anchor=base},baseline=0pt]
		\node[inner sep=20pt] (model) at (0,0.5) {\includegraphics[width=.4\textwidth]{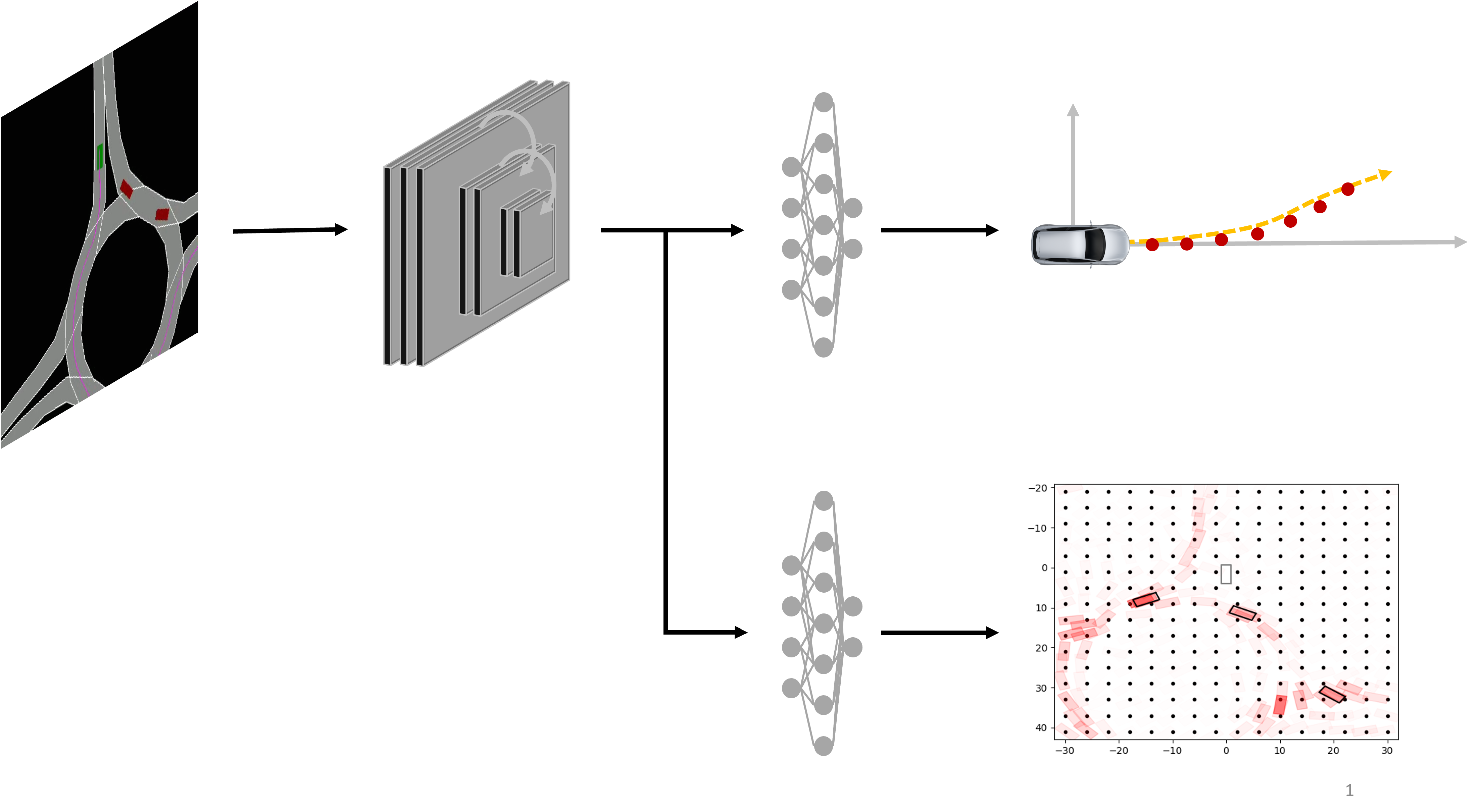}};
		\node[text width=50pt] (input) at (-3.2,5) {Binary grid\\features};
		\node[text width=50pt] (input) at (-3.2,2) {(as RGB)};
		\node[text width=40pt] (input) at (-1.3,4.7) {CNN\\Backbone};
		\node[text width=45pt] (input) at (0.4,4.7) {Waypoints\\Head};
		\node[text width=45pt] (input) at (2.4,4.7) {Waypoints\\Output};
		\node[text width=45pt] (input) at (0.4,0.4) {Auxiliary\\Head};
		\node[text width=45pt] (input) at (2.4,0.4) {Auxiliary\\Prediction};
	\end{tikzpicture}
	\caption{Model architecture consisting of a CNN backbone, a waypoints head, and a prediction head that generates the auxiliary prediction output.}
	\label{fig:model}
\end{figure}

For our experiments, we use a simple model architecture depicted in Fig.~\ref{fig:model} which is loosely inspired by~\cite{Bansal2018ChauffeurNet}.
As we focus on the training setup and development process instead of model architecture,
we opt for simple and well-established building blocks.

\subsection{Input Representation}
Our model consumes a list of dynamic traffic objects, a high-definition~(HD) map, and the current localisation of the self-driving vehicle~(SDV) on the map.
These inputs are rendered into a binary birds-eye-view grid of shape $[256, 256, 13]$ 
with a resolution of $0.25\times 0.25\ \mathrm{m}^2$ per pixel in the local frame of the SDV.
The SDV is positioned facing downwards at 84 pixels from top and 128 pixels from the left to allow for a slightly larger field of view to the front.
Of the 13 binary channels, 3 are reserved for SDV current and historic poses (separated by 200 milliseconds) and similar for dynamic traffic objects.
The remaining 7 channels contain two types of lane boundaries (solid or broken), drivable road surface, three speed limits, and the navigation route.

\subsection{Output Representation}
The model generates a control trajectory in the form of waypoints and an auxiliary prediction task.
The waypoint output contains $n$ pairs of $(x,y)$ coordinates with a fixed $\delta t$, hence, they are equidistant in time.
Based on empirical observations, $n=15$ and $\delta t=200$ milliseconds turned out to balance well between plan consistency and look-ahead.
The auxiliary prediction output consists of $16 \times 16$ anchors (same field of view as input grid).
For each anchor, object existence score, offset, and object orientation are regressed.
In this way, the model learns possible future evolutions of the scene.

\subsection{Network Architecture}
We employ a $\mathrm{ResNet}$-34 architecture, truncated after the second layer, as a backbone.
As default implementations expect RGB image inputs, the binary input grids are pre-processed by a $1\times 1$ convolutional layer.

The latent output is consumed by a waypoints head and an auxiliary prediction head.
The waypoints head is implemented by a convolutional layer with stride 2, kernel size 4, and 32 output channels,
followed by a two-layer MLP with hidden dimension of 128 that directly outputs the full control trajectory in the form of waypoints.

The prediction head is modelled after~\cite{redmon2016you} and consists of two $1\times 1$ convolutional layers with hidden dimension of 32.
We do not use larger convolutional kernels to enforce that future object locations are already encoded in the backbone.
In this way, this information is also available to the waypoints head.

\subsection{Trajectory Controller}
To generate a drivable trajectory based on the policy output, we first fit cubic splines to the waypoints
and then sample a trajectory including position, heading, and velocity.
We use separate linear controllers for acceleration and steering, both with control timepoint 400 milliseconds ahead.

For deployment in the test vehicle, we employ a different trajectory generator, which first samples drivable trajectories around the reference path and velocity profile defined by the waypoints.
The final trajectory is then chosen by an optimisation algorithm including standard comfort and safety costs.
Since the sampling width is parametrised to be very narrow, it acts as a smoother for the trajectory defined by the waypoints.
For low-level control in the car, we  apply a trajectory-follow controller which essentially consists of a PID control for longitudinal and lateral movement, respectively.

\section{Training Setup}
\label{training}

\subsection{Data Generation}
\label{training_data}
The bulk of training data is generated with the lightweight open-source simulator \highwayenv~\cite{highway-env}.
Both the ego agent and surrounding traffic are controlled by the built-in IDM model~\cite{treiber2000congested}.
We extended the built-in conflict-checking logic for a smoother and more realistic driving behaviour in roundabout scenarios.
For simulation, we relied on real-world maps of three roundabouts as well as rural roads.

The full dataset consists of approximately 3,000 simulated sequences resulting in around 730k single frames.
It can be generated within hours on consumer hardware.
For training, we are not using any recorded or real-world sequences.
The full dataset is split into training and development sets on sequence basis with 80--20 ratio.
Both splits are randomly sub-sampled until they contain around 240k and 37k frames, respectively.
See Table \ref{tab:scenarios_description} for a detailed list of the specific scenarios used for training.
We use the term scenario to refer to specific driving situations or environments, encompassing factors such as road layout, traffic conditions, and dynamic interactions with other road users.

\begin{table*}[t]
\centering
\begin{tabular}{|l|l|l|}
\hline
\textbf{Scenario} & \textbf{Description} & \textbf{Expected Behaviour} \\
\hline
Slow down behind vehicle & The SDV is driving faster than the vehicle ahead, and needs to reduce speed. & increase gap \\
\hline
Increase gap & Starting with an insufficiently small gap, the SDV must restore a safe distance. & increase gap \\
\hline
Stop behind vehicle & A vehicle blocks the SDV's lane for no apparent reason, the SDV needs to stop. & stop \\
\hline
%Very slow following & Following a vehicle moving particularly slowly is a scenario for the policy. \\
%\hline
Driveoff and stop behind vehicle & The SDV is waiting in line and the vehicle ahead moves forward a couple of metres. & stop \\
\hline
%Long-tail stopping events & Encountering atypical situations where stopping within a roundabout due to rare traffic conditions is necessary for the policy. \\
%\hline
Yield & The SDV must yield to crossing traffic before entering the roundabout. & go to goal \\
\hline
Yielding with preceding vehicle & The vehicle ahead of the SDV yields to crossing traffic before entering the roundabout. & go to goal\\
\hline
Random traffic & Roundabout scenario with varying traffic densities and patterns. & go to goal \\
\hline
%Static offroad vars & The policy is required to handle scenarios involving parked cars off the roadway, ensuring appropriate non-engagement. \\
%\hline
%No static offroad cars & The policy is required to handle scenarios involving no parked cars off the roadway. \\
%\hline
\end{tabular}
\caption{Description of scenarios used for data generation}
\label{tab:scenarios_description}
\end{table*}

\subsection{Augmentations}
\label{training_augmentations}

Via our training and evaluation iterations, we arrived at the set of the following data augmentations.

\subsubsection{Computer-Vision-Based Augmentations}
We apply random rotation of the input grid with the SDV in the rotation centre.
The rotation angle is uniformly distributed between $\pm 5^\circ$.
Additionally, we apply random translation of all channels of the input grid except the SDV channel,
uniformly distributed between $\pm 2$ metres.
Both augmentations are independently applied with probability $0.5$ each,
and labels adjusted accordingly.

\subsubsection{Off-Road Vehicles}
After observing that our trained policies are vulnerable to detected vehicles on a nearby parking lot,
we augment the simulated training sequences with randomly placed standing vehicles.
We explore two methods:
Firstly, randomly place vehicles in the vicinity of the road and fix for the full sequence.
Secondly, copy off-road vehicles from real-life data on frame-by-frame basis with SDV position matching,
capturing perception noise.
Empirically, we find that the former is sufficient to achieve robustness in our test scenarios.

\subsubsection{Traffic Vehicle Size}
As the size of different vehicles varies a lot in reality,
we randomly sample traffic object dimensions in simulation.
Length and width are picked uniformly and independently for each vehicle in $[4,6]$ and $[1,3]$ metres, respectively.

\subsubsection{Bitflip Augmentation}
There remains a sim-to-real gap besides the effects that have been addressed by the previously listed data augmentations,
e.g.\ track losses, false-positive detections, or general perception ``noise'' away from the region of interest.
We capture some of these effects with a generic noise term,
i.e.\ by randomly flipping single input grid pixels independently with probability $p_\textrm{flip}$.
We did not observe additional benefits from more elaborate schemes,
such as manipulation of object boundaries,
or flipping connected clusters of pixels.

\subsection{Model Training}
\label{training_training}
The full training loss reads
\begin{align}
\mathcal{L}
= \lambda_1 \cdot \mathcal{L}_\textrm{pos}^\textrm{plan}
+ \lambda_2 \cdot \mathcal{L}_\textrm{vel}^\textrm{plan}
+ \lambda_3 \cdot \mathcal{L}_\textrm{class}^\textrm{pred}
+ \lambda_4 \cdot \mathcal{L}_\textrm{reg}^\textrm{pred} \,.
\end{align}
The first two terms describe the planning loss,
where we apply smooth L1 (Huber) loss on the waypoints ($\mathcal{L}_\textrm{pos}^\textrm{plan}$)
and L2 loss on the differences between successive waypoints which can be interpreted as velocity ($\mathcal{L}_\textrm{vel}^\textrm{plan}$).
The latter two terms form the prediction loss,
focal loss~\cite{lin2017focal} with $\alpha=0.5$, $\gamma=2$ for classification ($\mathcal{L}_\textrm{class}^\textrm{pred}$),
and L1 loss for regression ($\mathcal{L}_\textrm{reg}^\textrm{pred}$).
The weights are $\lambda_1=0.5$, $\lambda_2=1$, $\lambda_3=10$, $\lambda_4=0.2$.

We use the Adam optimizer with initial learning rate $0.001$,
reduced by $0.94$ after each epoch (exponential learning rate scheduling),
weight decay regularization,
and batch size $256$.
All models are trained for 50 epochs.

\section{Evaluation}
\label{evaluation}

\subsection{Open-Loop Evaluation}
\label{evaluation_ol}
We measure imitation performance of trained policies w.r.t.~recorded data.
To make results more expressive, short snippets of
1-5 seconds length with a specific driving situation are labelled and extracted from longer sequences.
Specific scenarios include stopping behind a vehicle, keep standing in traffic, drive-off, following a preceding vehicle, yielding at an intersection, etc.

Auxiliary metrics are
``waypoints' inverse time-to-collision'', defined by the earliest waypoint in the plan that intersects with traffic (at planning time),
and classification for standstill, defined by two consecutive waypoints being closer than a threshold (10 centimetres which translates to 0.5 metres per second).

In addition, we define temporal plan instability as the deviation between (adjusted) planned trajectory end-points for consequtive timesteps
\begin{align}
\mathrm{tpi}(\tau) = \| \left(x,y\right)_T^{@ \tau} - \left(x,y\right)_{T-1}^{@ \tau+1} \|_2 \,.
\end{align}
$\left(x,y\right)_t^{@ \tau}$ is the planned waypoint $t$ seconds into the future, generated at time $\tau$.
A low value indicates that plans only change gradually over time and serves as a proxy for policy robustness and driving comfort.

\subsection{Closed-Loop Evaluation}
\label{evaluation_cl}
For closed-loop evaluation we rely on the same \highwayenv simulator that is used for data generation.
We define several logical scenarios centred around the roundabout use case with different
positioning and behaviour (e.g.\ set speed) of all traffic participants in the scene.
SDV navigation goal and the dimensions of traffic vehicles are sampled randomly.
The simulation terminates on crashes, when the SDV leaves the drivable area by more than 5 metres, or if the success criterion is fulfilled.

Scenarios can be grouped as follows according to different success criteria:
\begin{itemize}
\item{\textit{Go-to-goal scenarios:}
The SDV must navigate through the roundabout to a defined goal location.}
\item{\textit{Stop scenarios:}
Stopped vehicles block the lane ahead of the SDV, and it must stop in-lane.}
\item{\textit{Increase gap scenarios:}
The SDV is initialised in short distance behind a slower lead vehicle and must restore a safe follow distance.}
\end{itemize}

A listing the scenarios, along with their explanations, is presented in Table \ref{tab:scenarios_description}.
\subsection{Test Vehicle Integration}
\label{evaluation_vehicle}
\begin{figure}
	\centering
	\includegraphics[width=.3\textwidth]{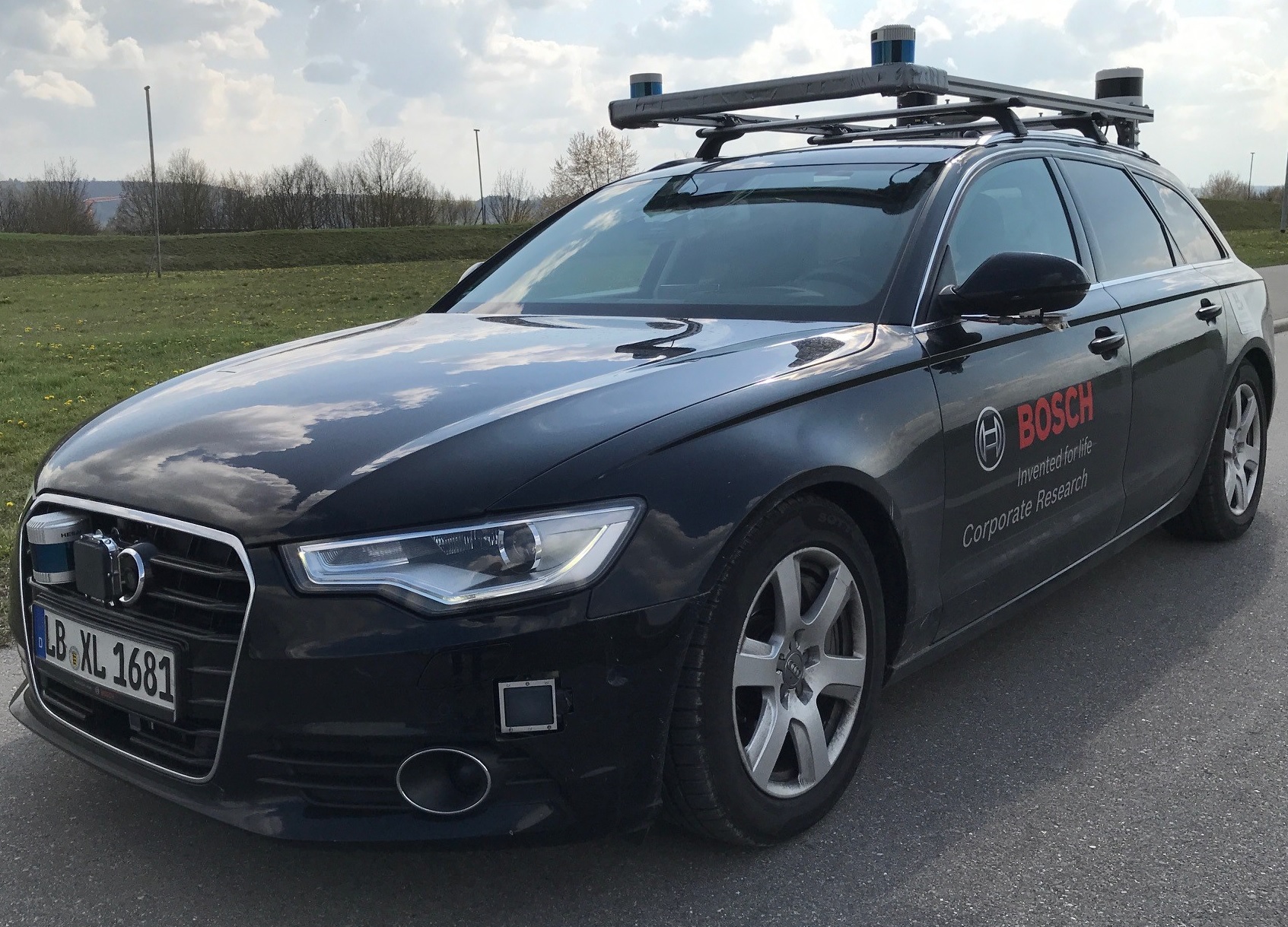}
	\caption{Test vehicle equipped with a prototypic autonomous driving stack.}
	\label{fig:test_vehicle}
\end{figure}

For real-world evaluation, we integrate our policy into a test vehicle equipped with a prototypic autonomous driving stack based on ROS (see Fig.~\ref{fig:test_vehicle}).
A list of tracked traffic participants is published at 10 Hertz, and SDV localisation w.r.t.\ the HD map provides motion data at 50 Hertz.
The trained policy is implemented as a ROS node,
and a trajectory smoother ensures drivability and comfort of the generated plan.
Safety checks could be easily added to the smoother which we omit to expose failures of our driving policy.

In addition to the basic functionalities, the following extensions are implemented:
\begin{itemize}
	\item \textit{Input frequency:}
	The policy requires past observations with temporal offsets of exactly 200 milliseconds.
	We therefore store received messages in a queue and then compensate for jitter to match the appropriate timestamps.
	\item \textit{SDV reference point:}
	The reference point of the SDV slightly differs in our training data (centre of gravity) and in our test vehicle (centre of rear axle).
	The transformation of the inputs is straightforward;
	we post-process the policy output by moving the points along the interpolated (raw) waypoints output.
	\item \textit{Cope with perception noise:}
	To cope with most of the perception noise (e.g.\ false positives, noisy object detection), we augment the training data as described in Section~\ref{training_augmentations}.
	The dimensions of traffic objects are set the most recent estimate and kept static over historical input timesteps.
\end{itemize}

\section{Results}
\label{results}
\subsection{Overall Driving Performance}
\label{results_overall}
The learned policy has been tested during multiple demo drives through roundabouts and country roads around Renningen, Germany.
Driving comfort was generally rated smooth.
In random situations that tend to have few interactions with traffic, the vehicle was able to safely navigate to the goal in $> 9/10$ drives.
In staged situations with additional lead vehicles and / or priority vehicles in the roundabout, driving without human intervention was achieved in $\sim 9/10$ drives,
which is reasonable given the absence of any safety module and the simplicity of the approach.

Note that this performance has only been reached after a series of alternating off-line and in-vehicle evaluations and a subsequent extensions of our training data.
Off-line performance (closed-loop or open-loop) is only a good proxy metric for real-world performance if the simulation is complete w.r.t.\ the variations found on the road.
Notable examples where our policy failed only in real-world tests until training data (and evaluation scenarios) were augmented are off-road vehicles on a nearby parking lot and variations in the size of vehicles.
In addition, we massively up-sampled the crucial interactions in roundabout driving by targeted simulation:
yielding to crossing traffic, additional preceding vehicles, traffic jam, and recovering a safe distance ahead.
We conclude that a simplistic simulator can be a sufficiently good proxy for real-world driving, if the gap between the two environments is regularly evaluated and the simulator extended accordingly.

With regard to driving comfort (smoothness and perceived safe distance ahead) we found that increased input grid resolution is superior, even if an automated hyperparameter search based on performance in simulation indicated that a much coarser resolution (by a factor of 4) would be sufficient.

Please note that our focus has been put on the approach for leveraging simulated data for scaling automated driving and a much better performance could be achieved with a more sophisticated planning model.

\subsection{Targeted Simulated Training Data}
A key feature of our approach is to massively upsample certain interactive scenarios that are typically underrepresented in datasets.
We found that it is sufficient to generate this data resource-efficiently with a simple simulator and a simplistic phenomenological modeling of perception effects such as tracking noise or track losses.
Some exemplary results when one of our synthetic scenarios is removed from the training set:
\begin{itemize}
\item \textit{Approaching the roundabout with an ahead vehicle:}
This scenario is fairly simple, because it is typically sufficient to keep a safe distance ahead to avoid collisions.
Success rate (i.e.\ leaving the roundabout on the correct lane withouth infractions) is 1.0, but drops to 0.98 due to collisions if those scenarios are not explicitly added to the train set.
\item \textit{Stopping Scenario:}
In this scenario, a vehicle blocks the SDV's lane for no apparent reason.
This is rarely seen in realistic datasets, but it is absolutely crucial that the SDV behaves safely.
If this scenario is removed from the train set, collision rates in the same scenario surge from 0.14 to 0.70, and even off-road driving rates increase from 0.13 to 0.25.
The effect is also seen when evaluating in a random scenario:
Collision rates increase from 0.07 to 0.11 (numbers for scenario with randomly parked cars).
\end{itemize}

\subsection{Comparison with Large Public Dataset}
\label{results_nuscenes}

\begin{table*}[t]
	\centering
	\begin{tabular}{ l | c c c | c c c}
		& \multicolumn{3}{c}{Random Scenario} & \multicolumn{3}{c}{Challenging \highwayenv Stopping Scenario} \\
		train set & Success $\uparrow$ & Collisions $\downarrow$ & Off-road $\downarrow$ & Success $\uparrow$ & Collisions $\downarrow$ & Off-road $\downarrow$ \\
		\hline
		\nuscenes              & 0.00 & 0.36 & 0.72 & 0.00 & 0.66 & 0.34 \\
		\nuscenes (subsampled) & 0.00 & $\mathbf{0.00}$ & 0.68 & 0.00 & $\mathbf{0.12}$ & 0.88 \\
		ours & $\mathbf{0.95}$ & 0.03 & $\mathbf{0.02}$ & $\mathbf{0.73}$ & 0.14 & $\mathbf{0.14}$ \\
	\end{tabular}
	\caption{Comparison of the network trained on \nuscenes and simplistic simulated data (ours), both evaluated in the simulator described in Section~\ref{evaluation_cl}.}
	\label{tab:nuscenes}
\end{table*}

For comparison we trained the same model on the \nuscenes public dataset~\cite{caesar2020nuscenes},
which covers a range of diverse driving situations from different locations.
The full train split contains approx.\ 93k single frames after pre-processing, and we used the data augmentation methods described in Section~\ref{training_augmentations}.
Two models were trained until convergence:
the first on all frames in the dataset, the second on randomly sampled 40\% of frames (same setup as our model).
Although the training set is much more diverse than our simplistic simulated data,
Table~\ref{tab:nuscenes} shows that policies trained on \nuscenes alone completely fail in the \highwayenv roundabout scenario.
Dominant failure cases differ for the two policies despite the nearly identical training setup, which is not uncommon for models that fail to generalise to the test cases.

Observed failure cases include:
\begin{itemize}
\item \textit{Failure to adjust speed:}
In some cases the policy speeds up in front of the roundabout, rendering lane keeping impossible.
In scenarios with an ahead vehicle, front crashes are common.
We note that these failure cases also appeared during our development, and could be solved by adding similar situations to the train set.
\item \textit{Standstill and freeze:}
It happens that the policy stops for cross traffic, or for no apparent reason, and fails to drive off again.
This hints at a the well-known problem of action repeat~\cite{muller2005off, codevilla2019exploring}.
\item \textit{Failure to stay on lane:}
Even without obstacles on the road, in some cases the policy leaves the lane.
The same problem could be solved by simple computer-vision-based augmentations for our policy, but seems to require a different setup for the \nuscenes dataset.
\end{itemize}

In essence, we confirm that a realistic dataset alone is insufficient to train a planning policy,
as already observed by e.g.~\cite{Bansal2018ChauffeurNet, scheel2022urban}.

\subsection{Ablation Studies}
\label{results_ablations}
Unless stated otherwise, ablation studies are based on closed-loop simulation as described in Section~\ref{evaluation_cl}.
Scenarios are chosen based on the largest impact of the ablation, comprising randomly generated scenarios at roundabouts with or without additionally added parked cars, challenging stopping scenarios and high-speed scenarios on rural roads.
The performance is then evaluated based on success, collision and off-road rates, as well as average speed over the entire scenario.
Success is defined by infraction-free driving along the navigation route.
Stopping scenarios are rated successful if the SDV comes to a complete stop in-lane.
Each scenario is executed 100-200 times per configuration, and we quote mean values.

\begin{table*}[t]
	\centering
	\begin{tabular}{ c c | c c c | c c c}
		&& \multicolumn{3}{c}{Random Scenario} & \multicolumn{3}{c}{Challenging Stopping Scenario} \\
		$\mathcal{A}_\textrm{rot}$ & $\mathcal{A}_\textrm{trans}$ & Success $\uparrow$ & Collisions $\downarrow$ & Off-road $\downarrow$ & Success $\uparrow$ & Collisions $\downarrow$ & Off-road $\downarrow$ \\
		\hline
		\xmark & \xmark & 0.87 & 0.04 & 0.09 & 0.57 & 0.28 & 0.17 \\
		\cmark & \xmark & 0.93 & 0.05 & $\mathbf{0.02}$ & 0.67 & $\mathbf{0.11}$ & 0.21  \\
		\xmark & \cmark & 0.79 & 0.06 & 0.15 & 0.63 & 0.17 & 0.22 \\
		\cmark & \cmark & $\mathbf{0.95}$ & $\mathbf{0.03}$ & $\mathbf{0.02}$ & $\mathbf{0.73}$ & 0.14 & $\mathbf{0.14}$ \\
	\end{tabular}
	\caption{Ablation study regarding rotational $\mathcal{A}_\textrm{rot}$ and translational $\mathcal{A}_\textrm{trans}$ augmentations as described in Section~\ref{training_augmentations}.}
	\label{tab:ablation_augmentation}
\end{table*}

\begin{table*}[t]
	\centering
	\begin{tabular}{ c c | c c c c | c c c c}
		&& \multicolumn{4}{c}{Random Scenario With Randomly Parked Cars} & \multicolumn{4}{c}{High-Speed Scenario} \\
		$\mathcal{A}_\textrm{dim}$ & $\mathcal{A}_\textrm{offroad}$ & Success $\uparrow$ & Collisions $\downarrow$ & Off-road $\downarrow$ & Speed[m/s] $\uparrow$ & Success $\uparrow$ & Collisions $\downarrow$ & Off-road $\downarrow$ & Speed[m/s] $\uparrow$ \\
		\hline
		\cmark & \xmark & 0.77 & 0.07 & 0.15 & 4.68 & 0.02 & 0.27 & $\mathbf{0.00}$ & 3.68 \\
		\xmark & \cmark & 0.84 & 0.08 & 0.06 & 5.51 & 0.71 & 0.01 & $\mathbf{0.00}$ & 6.93 \\
		\cmark & \cmark & $\mathbf{0.89}$ & $\mathbf{0.06}$ & $\mathbf{0.04}$ & $\mathbf{5.71}$ & $\mathbf{1.00}$ & $\mathbf{0.00}$ & $\mathbf{0.00}$ & $\mathbf{11.78}$ \\
	\end{tabular}
	\caption{Ablation study on the effect of augmenting off-road vehicles $\mathcal{A}_\textrm{offroad}$ and traffic dimensions $\mathcal{A}_\textrm{dim}$ as described in Section~\ref{training_augmentations}.}
	\label{tab:ablation_perturbation}
\end{table*}

\begin{table*}[t]
	\centering
	\begin{tabular}{ c c | c c c}
		$\mathcal{A}_\textrm{bitflip}$ & $p_\textrm{flip}$ & Stop-and-go & On priority lane & Yield \\
		\hline
		\xmark &      & 1.42 & 1.97 & 1.16 \\
		\cmark & 0.01 & 1.11 & 1.32 & $\mathbf{0.90}$ \\
		\cmark & 0.1  & $\mathbf{0.99}$ & $\mathbf{1.20}$ & 1.01 \\
	\end{tabular}
	\caption{Ablation study regarding bitflip augmentation as described in Section~\ref{training_augmentations}. Reported values are temporal plan instability, the difference between the end point of two concecutive plans in metres (cf.\ Section~\ref{evaluation_ol}), evaluated on recorded data.}
	\label{tab:ablation_bitflip}
\end{table*}

\begin{table*}[t]
	\centering
	\begin{tabular}{ c c | c c c | c c c}
		&& \multicolumn{3}{c}{Random Scenario} & \multicolumn{3}{c}{Challenging Stopping Scenario} \\
		$\mathcal{L}_\textrm{pos}^\textrm{plan}$ & $\mathcal{L}_\textrm{vel}^\textrm{plan}$  & Success $\uparrow$ & Collisions $\downarrow$ & Off-road $\downarrow$ & Success $\uparrow$ & Collisions $\downarrow$ & Off-road $\downarrow$ \\
		\hline
		\cmark & \xmark & 0.94 & $\mathbf{0.03}$ & 0.03 & 0.68 & 0.19 & 0.15 \\
		\xmark & \cmark & 0.00 & 0.00 & 1.00 & 0.00 & 0.00 & 1.00 \\
		\cmark & \cmark & $\mathbf{0.95}$ & $\mathbf{0.03}$ & $\mathbf{0.02}$ & $\mathbf{0.73}$ & $\mathbf{0.14}$ & $\mathbf{0.14}$ \\
	\end{tabular}
	\caption{Ablation study regarding planning losses as described in Section~\ref{training_training}.}
	\label{tab:ablation_loss}
\end{table*}

\textbf{Computer-vision-based augmentations:}
Table~\ref{tab:ablation_augmentation} shows the effect of adding computer-vision-based augmentations such as rotation $\mathcal{A}_\textrm{rot}$ and translation $\mathcal{A}_\textrm{trans}$.
We observe a significant improvement when adding rotational augmentations, especially to prevent off-road driving.
Translational augmentations slightly increase the success rate in challenging stopping scenarios.
However, with $\mathcal{A}_\textrm{trans}$, we observe a major improvement of the driving comfort in our test vehicle.

\textbf{Traffic vehicle augmentations:}
The effect of additional off-road vehicles $\mathcal{A}_\textrm{offroad}$ and perturbed dimensions of traffic participants $\mathcal{A}_\textrm{dim}$ is shown in Table~\ref{tab:ablation_perturbation}.
As expected, the off-road augmentation improves the policy performance in scenarios with many parked cars.
A major impact of both augmentations can be observed in scenarios with higher speed.
This is caused by faulty attentions to other traffic participants resulting in lower average speed.
Low success rates originate in the SDV not reaching the goal in-time.

\textbf{Bitflip augmentation:}
We found that randomly changing the binary value of pixels in the input grid during training as described in Section~\ref{training_augmentations} has no measurable effect in a clean simulation environment.
But on realistic inputs evaluated in open-loop, policies trained with said augmentation gave plans that are more stable over time across all evaluated scenarios.
Refer to Table~\ref{tab:ablation_bitflip} for exemplary values for temporal plan instability (recall the definition in Section~\ref{evaluation_ol}).

\textbf{Planning losses:}
Table~\ref{tab:ablation_loss} shows the impact of our planning losses $\mathcal{L}_\textrm{pos}^\textrm{plan}$ and $\mathcal{L}_\textrm{vel}^\textrm{plan}$.
As expected, the policy cannot be trained with velocity loss alone.
Adding velocity loss achieves only slight improvement in simulation,
but much better driving comfort in our test vehicle.
Experiments with different loss functions for $\mathcal{L}_\textrm{pos}^\textrm{plan}$ -- including separating longitudinal and lateral error, and dynamic loss weights -- showed negligible impact.

\subsection{Useful Off-Line Metrics}
\label{results_offline}
The relative value of different off-line evaluation metrics depends on the maturity and overall performance of the planner.
After having solved goal-conditioned driving on (almost) empty roads, we focused on safety in denser traffic.
As soon as we were satisfied with the real-world driving behaviour in the majority of scenarios and started to stage interactive situations with more traffic, we found the following off-line metrics most useful.

\textit{Closed-Loop Simulation:}
An important evaluation scenario that focuses on safe following is where the SDV must maintain a safe distance to a slowing vehicle ahead.
Distance-based metrics like inverse time-to-collision, violations of safety distance ahead, and collisions are good indicators of safe real-world driving.
Also scenarios where the SDV has to yield to priority traffic in the roundabout are helpful, if the constellations close to the point of conflict show enough variety in terms of relative time of arrival, and velocity.
Again distance-based metrics like inverse time-to-collision and collisions proved good proxy metrics.

\textit{Open-Loop Evaluation:}
With open-loop evaluation, it is not straightforward to define a reward for plans that deviate from the demonstrated (expert) trajectory.
Hence imitation-based metrics like $\mathrm{ADE}$ / $\mathrm{FDE}$ are often ill-suited.
Clearly defined ``atomic'' scenarios (see Section~\ref{evaluation_ol}) make it possible to define desired behaviour and thus help circumvent said problem.
For a focus on collision avoidance, we found metrics for waypoints' inverse time-to-collision and standing / stopping plans most useful in scenarios that require significant slowdown:
yielding to traffic, stopping behind a standing vehicle, and staying behind a standing vehicle.

For completeness, we list some open-loop metrics in Table~\ref{tab:openloop_eval_results}.

\begin{table*}[t]
	\centering
	\begin{tabular}{l | c c c}
		metric & yielding to traffic & stopping behind standing vehicle & staying behind standing vehicle\\
		\hline
		temporal plan instability & 0.36 & 1.21 & 0.61 \\
		waypoints inverse TTC & 0.00 & 0.50 & 0.22 \\
		waypoints standstill & 0.18 & 0.04 & 0.58 \\
		waypoints stopping & 0.25 & 0.10 & 0.55 \\
	\end{tabular}
	\caption{Selected open-loop evaluation results. Low temporal plan instability (cf.\ Section~\ref{evaluation_ol}) indicates smooth driving. Large inverse TTC and standstill / stopping rates are desirable in yielding or traffic jam-like scenarios.}
	\label{tab:openloop_eval_results}
\end{table*}

\section{Conclusions}
\label{conclusions}
This study challenges the widely-held belief that developing automated driving systems crucially needs huge amounts of real-world data or a highly realistic simulator.
In a roundabout scenario, we demonstrated the effectiveness to augment the training set with underrepresented driving situations using targeted, simplistic simulation data alone.

Utilising vanilla behavioural cloning and a simple planning model, we were able to achieve a comfortable real-world driving experience without relying on enourmous datasets.
Our iterative development process, which includes regular in-vehicle testing, has proven valuable for identification and prioritisation of sim-to-real gaps.
These gaps could be closed by data augmentation and scenario variations.
We expect that data aquisition and invest in more elaborate algorithms and architectures is becoming increasingly difficult for state-of-the-art planners,
and that a similar engineering effort is a viable and pragmatic alternative.

Overall, our study showcases the capability of a simple neural network architecture to handle various situations within the given scope of scenarios.
Unimodal model output proved sufficient here, and the problem of trajectory selection is circumvented.
This approach can therefore be integrated as an intelligent motion planner in a modular system, where an upstream module dictates the intended high-level behaviour.

\bibliographystyle{IEEEtran}
\bibliography{IEEEabrv,bibliography}

\end{document}